\PassOptionsToPackage{unicode}{hyperref}
\PassOptionsToPackage{hyphens}{url}
\PassOptionsToPackage{dvipsnames,svgnames,x11names}{xcolor}
\documentclass[
  10pt,
  letterpaper,
]{article}
\usepackage{xcolor}
\usepackage[margin=0.8in]{geometry}
\usepackage{amsmath,amssymb}
\usepackage{iftex}
\ifPDFTeX
  \usepackage[T1]{fontenc}
  \usepackage[utf8]{inputenc}
  \usepackage{textcomp} 
\else 
  \usepackage{unicode-math} 
  \defaultfontfeatures{Scale=MatchLowercase}
  \defaultfontfeatures[\rmfamily]{Ligatures=TeX,Scale=1}
\fi
\usepackage{lmodern}
\ifPDFTeX\else
\fi
\IfFileExists{upquote.sty}{\usepackage{upquote}}{}
\IfFileExists{microtype.sty}{
  \usepackage[]{microtype}
  \UseMicrotypeSet[protrusion]{basicmath} 
}{}
\makeatletter
\@ifundefined{KOMAClassName}{
  \IfFileExists{parskip.sty}{%
    \usepackage{parskip}
  }{
    \setlength{\parindent}{0pt}
    \setlength{\parskip}{6pt plus 2pt minus 1pt}}
}{
  \KOMAoptions{parskip=half}}
\makeatother
\usepackage{longtable,booktabs,array}
\usepackage{calc} 
\usepackage{etoolbox}
\makeatletter
\patchcmd\longtable{\par}{\if@noskipsec\mbox{}\fi\par}{}{}
\makeatother
\IfFileExists{footnotehyper.sty}{\usepackage{footnotehyper}}{\usepackage{footnote}}
\makesavenoteenv{longtable}
\usepackage{graphicx}
\makeatletter
\newsavebox\pandoc@box
\newcommand*\pandocbounded[1]{
  \sbox\pandoc@box{#1}%
  \Gscale@div\@tempa{\textheight}{\dimexpr\ht\pandoc@box+\dp\pandoc@box\relax}%
  \Gscale@div\@tempb{\linewidth}{\wd\pandoc@box}%
  \ifdim\@tempb\p@<\@tempa\p@\let\@tempa\@tempb\fi
  \ifdim\@tempa\p@<\p@\scalebox{\@tempa}{\usebox\pandoc@box}%
  \else\usebox{\pandoc@box}%
  \fi%
}
\def\fps@figure{htbp}
\makeatother
\providecommand{\tightlist}{%
  \setlength{\itemsep}{0pt}\setlength{\parskip}{0pt}}
\usepackage{bookmark}
\IfFileExists{xurl.sty}{\usepackage{xurl}}{} 
\hypersetup{
  pdftitle={Which Rules Matter Now? Policy-Centroid Routing Before an Intelligent System Acts},
  pdfauthor={Thomson D. Nguy},
  colorlinks=true,
  linkcolor={blue},
  filecolor={Maroon},
  citecolor={Blue},
  urlcolor={blue},
  pdfcreator={LaTeX via pandoc}}

\title{Which Rules Matter Now?\\[0.25em]
\large Policy-Centroid Routing Before an Intelligent System Acts}
\author{Thomson D. Nguy\\
\small Radiant Institute for Manifold Studies}
\date{}

\begin{document}
\maketitle

\begin{abstract}

Before an intelligent system can decide whether an action is allowed, it
must first know which rules the action has approached. A single proposed
action can implicate several policy regimes at once, and their
requirements can stack, overlap, or qualify one another. Many of those
rules remain expressed in natural language rather than encoded as
structured policy, while the action itself may arrive as an incomplete
description of intent. Before interpretation or enforcement can begin,
the system must determine which regimes belong in the review. The first
problem is not judgment. It is attention. {[}1, 2{]}

Policy-centroid routing creates a layer before adjudication: semantic
radar for determining which bodies of authority a proposed action has
approached. It treats the policy landscape as searchable geometry. The
expressions within each policy regime are compressed into one or more
representative centroids, and the proposed action is placed in the same
semantic space. A declared measure calculates proximity; a declared
threshold determines which regimes enter authoritative review. Several
regimes may trigger at once, carrying stacked or overlapping obligations
forward before the system acts.

If policy-centroid routing works, it would help AI agents and other
intelligent systems identify which policies may govern a proposed
action. Those policies may sit inside an overwhelming stack of
overlapping and sometimes conflicting obligations, many of which have
never been translated into deterministic rules. More of the rules that
matter could reach review before the system acts, while actual judgment
remains with the proper authority. The paper develops six falsifiable
claims and lays out seven follow-on studies to test them against
structured workflows, lexical and semantic retrieval, hierarchical and
direct classification, and selective prediction under the same review
burden. The studies are designed to show where policy geometry recovers
more applicable regimes, where it loses rare or overlapping obligations,
and where the mechanism should abstain. The synthetic worked example is
included in the paper, and a bounded reference implementation is publicly
available. The next step is to run the studies.


\end{abstract}

\section{1. The First Problem Is
Attention}\label{1-the-first-problem-is-attention}

An intelligent system often has to act before the institutional world
around the action has been fully named. A purchase can also transfer
data. A product change can trigger a disclosure. An ordinary message can
touch privacy, safety, employment, contract, and sector-specific rules
at once. The system cannot interpret every applicable rule until it
knows which rule systems have entered the picture. The first allocation
problem is simple to state: where should authoritative attention go?
{[}1, 2{]}

Formal policy architectures already solve the downstream version of this
problem. XACML distinguishes policy targets, applicable policies, policy
decisions, and enforcement. Given a structured decision request, it can
identify or retrieve the applicable policy components. By that point,
the request, its attributes, and the policy objects have already been
made explicit. {[}1{]}

The unresolved boundary appears before that structured request exists.
Open Policy Agent evaluates declarative policy against structured input;
an application or surrounding system must first supply the facts in
machine-readable form. Policy-text systems can annotate, classify, and
query long rules once a domain and taxonomy have been established. The
upstream question remains: how does an action described in the language
of the world become attached to the formal regimes that deserve
checking? {[}2, 4, 5{]}

The question can now be tested: can a compressed policy-level
representation route authoritative attention from an unstructured
proposed action toward the relevant policy regimes before adjudication
begins? We call the representation a policy centroid and the function
policy-centroid routing. The name carries both the mechanism and the
wager. {[}11, 14-19{]}

\section{2. Routing Before Judgment}\label{2-routing-before-judgment}

Applicability routing decides where to look. Its output is a set of
policy regimes: these are the bodies of authority that may govern what
you are about to do. A route may identify a broad domain, a policy
family, or another declared review target. It hands attention to
authority; it does not replace the examination that follows. {[}1{]}

Compliance adjudication takes over from there. It interprets the
applicable authority against the relevant facts, resolves conditions and
exceptions, and reaches whatever decision the governing institution
recognizes. Structured policy engines can perform parts of that work
when their rules and inputs are formalized. Legal and organizational
judgment may require additional evidence, authority, or human review.
Semantic proximity determines none of those outcomes: permission,
prohibition, legality, breach, notification duty, control
implementation, control effectiveness, or compliance. {[}1, 2{]}

Enforcement comes later still. In the XACML architecture, a policy
decision point evaluates a request and a policy enforcement point
carries out the resulting decision. Selection, evaluation, and
enforcement remain separate functions inside one governed process.
Policy-centroid routing stays on the selection side of that sequence. It
neither executes nor blocks the action. {[}1{]}

Routing can fail in several different ways. It can omit a regime that
should have been checked, trigger one that proves irrelevant, displace a
better route, or abstain because the evidence is insufficient. False
negatives, false positives, displacement, and abstention impose
different burdens. Selective-prediction research formalizes the tradeoff
between coverage and risk. Abstention has operational meaning only when
the system defines what authoritative fallback follows. {[}19{]}

Adjacent tasks look deceptively similar. Retrieval can surface a
statutory article without establishing that it governs. Question
answering can extract a sentence from a privacy policy without resolving
the proposed action. Policy-centroid routing succeeds only if its output
remains a routing object and every later judgment retains its own
authority. {[}1, 6, 7{]}

\section{3. Turning Policies Into Searchable
Geometry}\label{3-turning-policies-into-searchable-geometry}

Policy-centroid routing turns a field of policies into a searchable
geometry. The expressions belonging to each policy regime are encoded
and compressed into one or more representative centroids. A proposed
action is encoded in the same space. The system measures its proximity
to each centroid and routes every regime whose score crosses a declared
threshold.

\begin{center}
\includegraphics[width=0.58\linewidth,keepaspectratio,alt={Figure 1. Policy-centroid routing represented as distance in a shared semantic space.}]{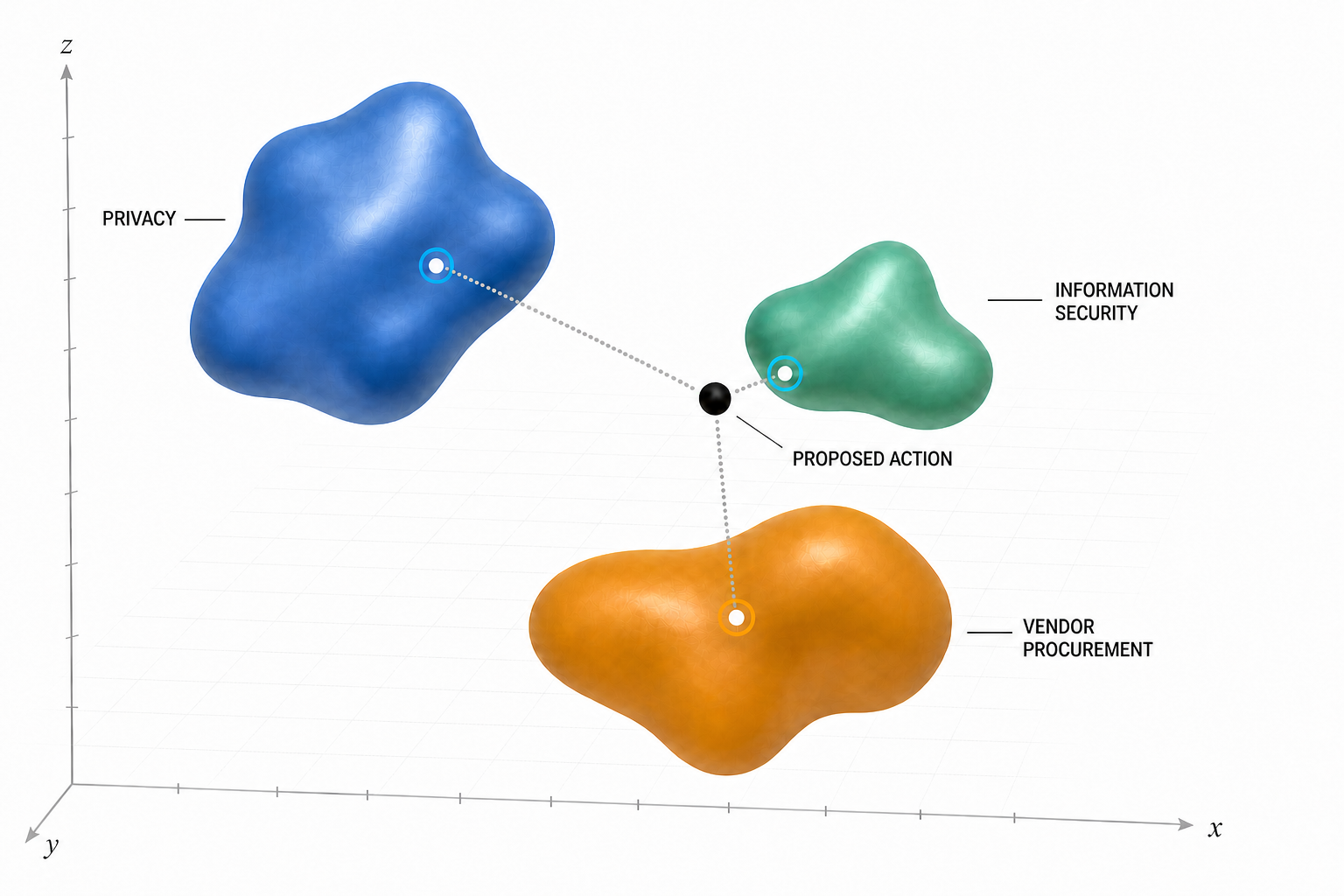}
\end{center}

\begingroup\small
\textbf{Figure 1. Policy-centroid routing as distance in a shared
semantic space.} This schematic projects a higher-dimensional
policy-embedding space onto three axes. The black sphere is a proposed
action; each colored field represents a policy regime and its marked
centroid. Dotted segments indicate distance to each centroid, not a path
through physical space. Here, information security lies nearest, vendor
procurement farther away, and privacy farthest. A declared measure and
threshold determine which regimes enter review. The geometry and
distances are illustrative; they are not empirical results or findings
of policy applicability.
\endgroup

The geometry depends on design choices. The encoder determines which
distinctions enter the space. The centroid construction determines what
gets compressed together. Cosine similarity, Mahalanobis distance, and
declared divergence measures each define proximity differently. A regime
can have one centroid or several. Representation and measure jointly
determine what the router can see. {[}11, 12{]}

\subsection{3.1 A Canonical Baseline}\label{31-a-canonical-baseline}

The hypothesis can be stated without leaving the baseline mechanism to
interpretation. Let

\[
\mathcal{P}=\{P_1,\ldots,P_m\}, \qquad
P_j=(\mathrm{id}_j,\ell_j,E_j)
\]

be a finite collection of policy regimes. Each regime has a unique
identifier, a reader-facing label, and a nonempty finite sequence of
policy expressions

\[
E_j=(p_{j1},\ldots,p_{jn_j}).
\]

A regime is a body of authority fixed by the institution before routing;
the geometry does not discover or redefine it. Expression assignment need
not be exclusive: the same language may inform more than one regime when
their authority overlaps.

Let \(\phi:\mathcal{T}\rightarrow\mathbb{R}^{d}\) be an encoder fixed and
declared before evaluation, mapping both policy text and proposed actions
into the same finite-dimensional space. Define the total normalization
operator

\[
N(v)=
\begin{cases}
v/\lVert v\rVert_2, & \lVert v\rVert_2>0,\\
0, & \lVert v\rVert_2=0.
\end{cases}
\]

The canonical construction is

\[
z_{ji}=N(\phi(p_{ji})), \qquad
\bar z_j=\frac{1}{n_j}\sum_{i=1}^{n_j}z_{ji}, \qquad
c_j=N(\bar z_j).
\]

Thus the canonical policy centroid \(c_j\) is the normalized arithmetic
mean of the regime's normalized expression vectors. For a nonempty proposed
action \(x\), let \(a=N(\phi(x))\). The baseline score is
cosine similarity,

\[
s_j(x)=a^{\mathsf T}c_j,
\]

and, for a finite declared threshold \(0<\tau\leq1\), the routing set is

\[
R_{\phi,\tau}(x)=\{(\mathrm{id}_j,\ell_j,s_j(x)):\ s_j(x)\geq\tau\}.
\]

The inequality is inclusive. Every crossing is returned; equal scores do
not require a tie-break for membership, and the router does not truncate
the set to a top-\(k\) list. Multiple regimes may therefore enter review
for one action. An empty routing set is an abstaining result requiring the
institution's declared fallback, never permission to act. Empty regimes,
duplicate identifiers, malformed text, dimension mismatches, and non-finite
vectors or arithmetic fail closed before a routing output. A zero action or
centroid vector receives no positive-threshold route rather than an invented
direction.

This definition is parameterized by \(\phi\) and \(\tau\), both of which
must be prespecified in an experiment. The synthetic reference instance
uses a deterministic TF--IDF encoder fitted only to policy expressions,
normalized arithmetic centroids, cosine similarity, and the inclusive rule
above so that the routing loop can be inspected. That encoder is not claimed
to be the best empirical representation. Alternative encoders, distance
measures, multiple-centroid constructions, and threshold-selection methods
are experimental variants, not silent degrees of freedom in this baseline.
The equations define an inspectable routing object; they provide no evidence
that its routes correspond to actual policy applicability.

The routing threshold and the review budget perform different jobs. The
threshold belongs inside the mechanism: it converts a proximity score
into a route. The review budget belongs to the evaluation: it gives
competing methods the same allowance and prevents any system from buying
recall by returning every policy. {[}19{]}

The mechanism stops after routing. Triggered regimes pass to
authoritative review, where the responsible institution interprets
conditions, exceptions, conflicts, and consequences. Policy-centroid
routing decides where attention should go before judgment begins.

\subsection{3.2 One Action, Several
Rulebooks}\label{32-one-action-several-rulebooks}

Suppose an employee proposes: "Send two weeks of customer-support
transcripts to a new outside AI vendor so it can test automated
summaries."

The sentence carries the business intent. It leaves out facts the
institution may need before the action can proceed. Where will the
vendor process the transcripts? Do they contain personal or restricted
data? Will the vendor retain them or use them to improve another model?
Has the vendor already passed procurement and security review? The
proposal may approach several bodies of authority before anyone can
answer those questions.

In a hypothetical routing pass, the proposed action is encoded once and
compared with the policy representations already constructed for the
institution. The output might look like this:

{\def\LTcaptype{none} 
\begin{longtable}[]{@{}>{\raggedright\arraybackslash}p{0.15\linewidth}>{\raggedright\arraybackslash}p{0.17\linewidth}>{\raggedright\arraybackslash}p{0.26\linewidth}>{\raggedright\arraybackslash}p{0.34\linewidth}@{}}
\toprule\noalign{}
Policy regime & Illustrative routing state & Why the action approached
it & Question left for authoritative review \\
\midrule\noalign{}
\endhead
\bottomrule\noalign{}
\endlastfoot
Vendor procurement & Routed & A new outside vendor will receive
institutional data & Has the vendor been approved, and what contractual
review is required? \\
Privacy and data handling & Routed & Customer transcripts may contain
personal information & What data are present, and is the proposed use
permitted? \\
Information security & Routed & Data would leave the
institution\textquotesingle s existing systems & What transfer, access,
storage, and deletion controls apply? \\
AI governance & Routed & The transcripts would be used to test an AI
system & What review applies to the model, purpose, and handling of its
outputs? \\
Cross-border transfer & Near the routing boundary & The
vendor\textquotesingle s processing location is missing & Does geography
introduce another governing regime? \\
Records retention & Not routed & The proposal says nothing about
temporary copies or deletion & Did compression hide an obligation that
should have entered review? \\
\end{longtable}
}

The route creates the agenda for review. Procurement, privacy, security,
and AI governance move forward together. Approval and prohibition remain
downstream decisions. The cross-border question exposes a missing fact.
Records retention stays outside the route unless a reviewer, another
system, or a better representation recovers it.

The same example shows how the mechanism can fail. A broad privacy
centroid could trigger whenever customer data appears, creating review
that adds no value. A narrow retention rule governing temporary test
copies could be averaged away. A cross-border restriction buried in
exceptional language might never pull the action across the threshold.
The scientific question is whether the geometry preserves more of the
governing stack under the same review burden, including the rare
obligations compression is most likely to erase.

\section{4. Policy Geometry as a Scientific
Hypothesis}\label{4-policy-geometry-as-a-scientific-hypothesis}

The wager is straightforward: actions and policy regimes may share
semantic structure even when they share little surface wording. If a
representation preserves that structure, distance in the representation
space may reveal governed territory that a literal match would rank
poorly. Prototype classification and dense, sparse-expansion, and
late-interaction retrieval show that learned representations can
organize other tasks. They supply plausibility for the geometry, not
evidence of policy applicability. That proposition remains untested.
{[}11, 15-17{]}

A serious test begins with structured workflows. XACML can select and
evaluate formal policy components against attributed requests. OPA
evaluates declarative rules against structured JSON input. OSCAL
provides machine-readable control documentation for validation,
conversion, and assessment workflows. Where facts and policy objects are
already formalized, these systems may be exact, auditable, and entirely
sufficient. A semantic front door earns its place only by solving the
upstream problem or adding measurable value under matched facts and
labor. {[}1, 2, 3{]}

Generic semantic retrieval may explain any apparent gain. A dense
retriever can embed an action and policy passages without constructing
policy centroids. A learned sparse method can expand beyond literal
vocabulary while retaining an indexable representation. A
late-interaction model can preserve token-level signals that a single
vector would compress. If any of them matches or exceeds the proposed
route under the same corpus and burden, generic retrieval is carrying
the result. {[}15-17{]}

Direct classification is another serious rival. Multilabel and
hierarchical classifiers can exploit known label structures when labeled
examples exist. A direct language model can classify text against an
explicit taxonomy, as Llama Guard demonstrates on the separate task of
conversational-safety classification. These methods may route as well as
or better than a centroid. Their data requirements, model access,
taxonomy assumptions, and inference budgets differ. A fair comparison
must account for those differences. {[}8-10, 18{]}

Coverage can counterfeit a geometry effect. A centroid may appear to
improve routing because it was built from longer, richer, or
better-matched policy material, because its corpus contains the relevant
language, or because its source unit makes recall easier. Representation
length and term weighting affect ranking; legal-retrieval results also
depend on the admitted corpus and answer universe. If the advantage
disappears when source material and information opportunity are matched,
geometry was not the cause. {[}7, 14, 15{]}

Compression can also defeat itself. A policy space becomes searchable by
throwing information away, and that discarded information may contain
the obligation that matters. Long-tail classification shows how
aggregate performance can conceal rare labels. Mixture-prototype work
shows that one representative may fit a complex class poorly.
Polythetic-classification theory shows the strain placed on prototype
thresholds when membership depends on combinations rather than one
dominant feature. These neighboring results make tail loss a prediction
to test, not an observed failure of policy centroids. {[}8, 12, 13{]}

\section{5. What the Theory Predicts}\label{5-what-the-theory-predicts}

\textbf{Proposition 1: comparative routing under matched burden.} Can a
declared centroid construction recover applicable policy regimes better
than the strongest feasible alternatives when policy information, input
facts, model opportunity, human labor, and review burden are matched?
The contest must include structured workflows where their prerequisites
are available, along with lexical, learned-sparse, dense, multi-vector,
supervised, and direct-model alternatives where scientifically
defensible. Ties and losses count. A win against a weak comparator does
not. {[}1, 2, 14-18{]}

Ranked relevance and complete applicable-set recovery answer different
questions. A top-ranked policy item can help while the returned set
still omits another regime governing the same action. Complete
applicability also depends on an oracle that may itself be incomplete.
Evaluation must report ranking behavior and recovery of independently
adjudicated applicable sets, including uncertainty and missingness.
Top-k retrieval asks which item ranks first. This paper asks whether
every applicable regime survived the route. {[}7, 8, 14-17{]}

\textbf{Proposition 2: early warning without the answer.} Can routing
information survive when the credited downstream rule language is absent
from the policy centroid? Policy question answering and statutory
retrieval begin with known documents or corpora and seek answer-bearing
text. This hypothesis concerns earlier warning at the policy-regime
level: can the representation detect governed territory before the
detailed answer has been supplied? It makes no claim of open-world
discovery or recovery of an unknown atomic rule. {[}6, 7, 15-17{]}

\textbf{Proposition 3: preservation of overlap and the tail.} Can the
route preserve every applicable regime when actions are multilabel,
including narrow or low-prevalence policy families? Legal
topic-classification datasets show how macro, rare-label,
hierarchy-depth, and exact-set views can diverge sharply from aggregate
scores. Those datasets provide no legal-applicability truth for this
paper. They do show how a system can look broadly competent while
failing where labels are sparse or overlapping. {[}8-10{]}

\textbf{Proposition 4: value under burden.} Does any recovery advantage
survive the cost of irrelevant routes, displaced correct routes,
indexing and inference, and hidden expert labor? A fixed-total
comparison asks what each method recovers under the same allowance. An
incremental comparison measures the added burden created by an
augmentation. Both are necessary because returning more material can
raise recall while making authoritative review less usable. The review
budget is an evaluation constraint, not a new allocator. {[}14-17, 19{]}

\textbf{Proposition 5: augmentation without a strawman.} Can a centroid
front door strengthen a serious conventional workflow? The miss set must
be fixed before anyone sees the centroid output. Conditional recovery
among those misses must be reported separately from value across the
full population. The structured workflow must receive the facts it was
designed to use; the semantic method cannot quietly receive richer
language or human interpretation. Otherwise the experiment measures
input asymmetry rather than augmentation. {[}1-3{]}

\textbf{Proposition 6: knowable boundaries.} Can the
method\textquotesingle s limits be described by domain, hierarchy layer,
policy version, language, action population, and abstention behavior?
MultiEURLEX documents temporal and granularity effects in a different
legal-classification task. Selective prediction supplies risk-coverage
vocabulary, and concept-drift synthesis distinguishes changing
distributions from static evaluation. These concepts can define bounded
evidence cells. One favorable threshold cannot be pooled into universal
readiness, and this paper proposes neither online adaptation nor a
generic account of policy revision as drift. {[}9, 19, 20{]}

\section{6. Where Compression Can
Fail}\label{6-where-compression-can-fail}

Arithmetic compression gives common language the most influence over the
representative. Stable central tendencies make a large class easier to
summarize, but policy significance does not follow frequency. A rare
exception, a low-prevalence duty, or a minority policy family may matter
more than the language dominating the centroid. Long-tail classification
and multimodal prototype research make this a concrete risk to test;
failure in policy centroids remains unobserved. {[}8, 12{]}

Multilabel collapse is the action-side version of the same danger. An
action may sit close to one dominant regime while another applicable
regime disappears below the threshold. Scoring ``any applicable route
found'' as success would reward partial recognition and conceal the
omission. Exact-set recovery, per-regime recall, and overlap-specific
slices are necessary because governance can fail through the missing
second answer. EUROVOC topics are labels rather than governing regimes,
but multilabel legal classification makes the measurement problem
visible. {[}8-10{]}

Conditional rules create a geometry with no dominant feature.
Applicability may depend on a conjunction: an actor of a certain type,
handling a certain object, in a certain jurisdiction, for a particular
purpose, above or below a threshold. Polythetic-classification theory
explains why prototype thresholds can struggle when combinations define
membership. Its consequence for policy routing remains an empirical
question. {[}13{]}

Every geometry embeds choices. The encoder determines which distinctions
enter the space. The policy unit determines what gets compressed
together. The measure defines ``near,'' and the threshold turns
distances into routes. Dense single-vector, learned sparse, and
late-interaction systems preserve different information at different
costs. Representation and metric choices must be prespecified. Choosing
the geometry after seeing the outcomes turns a test into a search for a
win. {[}11, 15-17{]}

The oracle can fail before the router does. PolicyQA operates within
selected policy documents. BSARD is bounded to Belgian legal questions
and a corpus of statutory articles; some questions require unavailable
or non-statutory sources. An incomplete oracle can punish a route for
finding a plausible regime the annotation never represented, or reward
the system for ignoring authority outside the corpus. Unknown and
unavailable must remain distinct from negative. {[}6, 7{]}

Staleness hides several events under one word. A policy may be amended,
changing the normative object. The corpus may change while the action
population remains stable. The population may change while the policy
stays fixed. The statistical relationship between action features and
labels may drift. MultiEURLEX uses chronological splits to expose
temporal effects in legal topic classification, while concept-drift
research supplies a broader statistical taxonomy. A future bounded study
may compare frozen policy versions with full recomputation. It must name
which kind of change it is testing; this paper proposes no online
updater. {[}9, 20{]}

\section{7. A Program of Decisive
Studies}\label{7-a-program-of-decisive-studies}

The theory produces seven follow-on studies. Each answers a distinct
empirical question, and evidence from one cannot fill an untested cell
in another. The studies remain proposals: each still requires its own
specification, review, execution, verification, and report.

Study 1 asks whether the simplest declared construction merits further
study as a broad applicability router against serious alternatives at
matched burden. Study 2 looks for the place where compression destroys
rare, conditional, minority, exceptional, or overlapping structure even
when the aggregate looks favorable. Study 3 removes downstream
answer-bearing language and its ancestry, then asks whether any signal
remains. Together they separate broad routing effects from tail harm and
information leakage. {[}8, 11-18{]}

Study 4 places policy-centroid routing beside a serious structured
workflow. It separates recovery among prospectively fixed workflow
misses from value across the full action population. The accounting
includes the adapter and human labor needed to turn an unstructured
action into the facts a conventional system expects. A conditional gain
that vanishes under full-population or fixed-total accounting narrows
the contribution. {[}1-3{]}

Study 5 asks a conditional handoff question: where defensible policy
hierarchies exist, does a broad centroid front door improve downstream
review-target recovery when paired with a conventional resolver?
Hierarchy traversal, resolver design, calibration, and budget allocation
sit outside this paper, so any implementation requires independent
specification and review. The study names the dependency without
smuggling in a solution. {[}9, 10{]}

Study 6 tracks a prespecified construction across frozen policy
revisions using full recomputation. Study 7 asks whether a prespecified
mechanism and its failure surfaces replicate in a separately evaluated
domain. A replication cannot carry its threshold, taxonomy, oracle, or
claim ceiling across jurisdictions. Each new domain must establish those
elements for itself. {[}9, 20{]}

The reporting rule is simple: show the complete disposition. The
mechanism may win, tie, harm, abstain, become infeasible, or become
uninterpretable because the oracle or source contract cannot support the
question. Rare-policy and overlap harms remain visible beside a
favorable aggregate. Abstention remains visible beside coverage. A null
result bounds the theory. A failure tells us where compression has
ceased to preserve the governed world. {[}8, 19{]}

\section{8. What Is Known and What Remains
Untested}\label{8-what-is-known-and-what-remains-untested}

Established work supplies the pieces around this problem. Formal policy
architectures select and evaluate rules over structured requests.
Policy-as-code systems execute declarative decisions against structured
input. Policy-text systems annotate, classify, query, and retrieve from
bounded corpora. Legal classification is multilabel, hierarchical,
long-tailed, multilingual, and temporally sensitive. Prototype methods
compress classes. Retrieval systems preserve lexical, vector, or
token-level structure. Direct models classify against supplied
taxonomies; selective systems abstain; drift research distinguishes
forms of change. Together these fields define the comparison set. None
has established policy-centroid routing. {[}1-20{]}

The mechanism examined here is deliberately narrow: represent a proposed
action as a vector; derive one or more policy centroids from policy
expressions; calculate a declared similarity, distance, or divergence;
apply a declared routing threshold; and hand potentially applicable
regimes to downstream review. The hypothesis does not extend to
downstream adjudication, enforcement, or the broader systems in which
such routing might operate.

The empirical ledger for this mechanism is still blank. This paper
contains no completed study, pilot result, null result, or operational
efficacy claim. The scientific question remains open.

The seven studies now carry the burden of answering the substantive
questions. Does the geometry improve applicability recovery? Does it
carry information beyond coverage and generic retrieval? Does it
preserve overlap and the tail? Does any gain survive burden accounting
or add value to structured workflows? Where does the mechanism stop
working? Evidence from this paper or one study cannot answer another
study\textquotesingle s question automatically.

\section{9. Toward Governed
Attention}\label{9-toward-governed-attention}

Governance begins by deciding where to look. Established policy systems
already separate functions inside formal architectures. Policy-centroid
routing asks whether a semantic layer can allocate attention earlier,
while the proposed action is still incomplete natural language and
before it has become a structured request. The layer routes, abstains,
and exposes uncertainty. The authority that owns the rules retains their
meaning and force. {[}1, 2, 19{]}

Success has a hard definition. ``More'' means more applicable regimes
under a fixed review burden, rather than a longer list purchased with
unbounded labor. ``Governed'' requires applicability, not merely
semantic resemblance. ``Territory'' must preserve stacked regimes
instead of collapsing them into the easiest answer. Policy geometry
earns its place only after structured workflows, strong retrieval,
supervised and direct classification, abstention, and
compression-pathology accounts have had their chance to win. {[}8,
11-19{]}

Compression is the promise and the danger. A good representation makes a
large policy space searchable. A bad one makes the inconvenient rules
disappear. The decisive question is simple: when does compression
preserve the rules that matter, and when does it make them disappear?
{[}8, 12, 13{]}

\section{Code and Reproducibility}\label{code-and-reproducibility}

A bounded reference implementation is available from the Seldon Reflex
Foundation at
\href{https://github.com/SRF-PBC/policy-centroid-routing}{github.com/SRF-PBC/policy-centroid-routing}.
It implements the declared routing loop, returns the complete ordered score
record, and fails closed on malformed or non-finite inputs. The deterministic
TF-IDF encoder and cosine measure make the mechanism inspectable; this is a
synthetic executable demonstration, not the empirical system proposed for
the seven studies and not evidence of policy-applicability performance.

\section{Epistemic Status}\label{epistemic-status}

{\def\LTcaptype{none} 
\begin{longtable}[]{@{}>{\raggedright\arraybackslash}p{0.36\linewidth}>{\raggedright\arraybackslash}p{0.58\linewidth}@{}}
\toprule\noalign{}
Manuscript surface & Status \\
\midrule\noalign{}
\endhead
\bottomrule\noalign{}
\endlastfoot
Attention-allocation problem and routing/adjudication distinction &
Theory and definition, grounded in prior formal-policy architecture \\
Canonical normalized-arithmetic-centroid/cosine/threshold routing baseline &
Proposed mechanism; no empirical result reported \\
Public bounded reference implementation & Executable demonstration at the
SRF repository; not an efficacy result \\
Policy-centroid efficacy & Untested \\
Geometry, coverage, workflow-sufficiency, generic-retrieval,
direct-classification, and compression-pathology accounts & Competing
theories and rivals \\
Studies 1 through 7 & Prospective, separately evaluated studies; none
executed or reported by this paper \\
Prior empirical evidence for this mechanism & None reported \\
\end{longtable}
}

\section{Disclosures}\label{disclosures}

\textbf{Conflict of interest.} The author is an inventor on patent
applications related to the concepts discussed in this manuscript.

\textbf{AI assistance.} The author originated and directed the research
question, conceptual framework, analysis, and conclusions presented in this
work. OpenAI Codex was used under the author's direction to assist with
literature organization, drafting, editing, and technical formatting. The
author independently verified the claims and citations, approved the final
text, and takes full responsibility for the work.

\section{References}\label{references}

\begin{enumerate}
\def\labelenumi{\arabic{enumi}.}
\tightlist
\item
  OASIS. \emph{eXtensible Access Control Markup Language (XACML) Version
  3.0}. OASIS Standard, 22 January 2013.
  \url{https://docs.oasis-open.org/xacml/3.0/xacml-3.0-core-spec-os-en.html}.
\item
  Open Policy Agent project. \emph{Open Policy Agent Documentation} and
  \emph{Integrating OPA}. Accessed 26 August 2026.
  \url{https://www.openpolicyagent.org/docs};
  \url{https://www.openpolicyagent.org/docs/integration}.
\item
  Piez, W. A. ``The Open Security Controls Assessment Language (OSCAL):
  Schema and Metaschema.'' \emph{Balisage: The Markup Conference} 2019.
  \url{https://doi.org/10.4242/BalisageVol23.Piez01}. Official NIST
  contribution published in the Balisage proceedings; the article states
  that the author\textquotesingle s opinions do not necessarily
  represent NIST.
\item
  Wilson, S., et al. ``The Creation and Analysis of a Website Privacy
  Policy Corpus.'' \emph{Proceedings of ACL}, 2016, 1330-1340.
  \url{https://doi.org/10.18653/v1/P16-1126}.
\item
  Harkous, H., et al. ``Polisis: Automated Analysis and Presentation of
  Privacy Policies Using Deep Learning.'' \emph{USENIX Security
  Symposium}, 2018, 531-548.
  \url{https://www.usenix.org/conference/usenixsecurity18/presentation/harkous}.
\item
  Ahmad, W. U., et al. ``PolicyQA: A Reading Comprehension Dataset for
  Privacy Policies.'' \emph{Findings of EMNLP}, 2020, 743-749.
  \url{https://doi.org/10.18653/v1/2020.findings-emnlp.66}.
\item
  Louis, A., and Spanakis, G. ``A Statutory Article Retrieval Dataset in
  French.'' \emph{Proceedings of ACL}, 2022, 6789-6803.
  \url{https://doi.org/10.18653/v1/2022.acl-long.468}.
\item
  Chalkidis, I., et al. ``Extreme Multi-Label Legal Text Classification:
  A Case Study in EU Legislation.'' \emph{Natural Legal Language
  Processing Workshop}, 2019, 78-87.
  \url{https://doi.org/10.18653/v1/W19-2209}.
\item
  Chalkidis, I., Fergadiotis, M., and Androutsopoulos, I. ``MultiEURLEX:
  A Multilingual and Multi-label Legal Document Classification Dataset
  for Zero-shot Cross-lingual Transfer.'' \emph{EMNLP}, 2021, 6974-6996.
  \url{https://doi.org/10.18653/v1/2021.emnlp-main.559}.
\item
  Zhou, J., et al. ``Hierarchy-Aware Global Model for Hierarchical Text
  Classification.'' \emph{Proceedings of ACL}, 2020, 1106-1117.
  \url{https://doi.org/10.18653/v1/2020.acl-main.104}.
\item
  Snell, J., Swersky, K., and Zemel, R. ``Prototypical Networks for
  Few-shot Learning.'' \emph{NeurIPS}, 2017.
  \url{https://proceedings.neurips.cc/paper/6996-prototypical-networks-for-few-shot-learning}.
\item
  Allen, K. R., et al. ``Infinite Mixture Prototypes for Few-shot
  Learning.'' \emph{ICML}, 2019, 232-241.
  \url{https://proceedings.mlr.press/v97/allen19b.html}.
\item
  Day, B. J., et al. ``Attentional Meta-learners for Few-shot Polythetic
  Classification.'' \emph{ICML}, 2022, 4867-4889.
  \url{https://proceedings.mlr.press/v162/day22a.html}.
\item
  Robertson, S., and Zaragoza, H. ``The Probabilistic Relevance
  Framework: BM25 and Beyond.'' \emph{Foundations and Trends in
  Information Retrieval} 3(4), 2009, 333-389.
  \url{https://doi.org/10.1561/1500000019}. Peer-reviewed scholarly
  synthesis; used here for bounded field vocabulary and formulation, not
  original empirical or priority claims.
\item
  Karpukhin, V., et al. ``Dense Passage Retrieval for Open-Domain
  Question Answering.'' \emph{EMNLP}, 2020, 6769-6781.
  \url{https://doi.org/10.18653/v1/2020.emnlp-main.550}.
\item
  Formal, T., Piwowarski, B., and Clinchant, S. ``SPLADE: Sparse Lexical
  and Expansion Model for First Stage Ranking.'' \emph{SIGIR}, 2021,
  2288-2292. \url{https://doi.org/10.1145/3404835.3463098}.
\item
  Khattab, O., and Zaharia, M. ``ColBERT: Efficient and Effective
  Passage Search via Contextualized Late Interaction over BERT.''
  \emph{SIGIR}, 2020, 39-48.
  \url{https://doi.org/10.1145/3397271.3401075}.
\item
  Inan, H., et al. ``Llama Guard: LLM-based Input-Output Safeguard for
  Human-AI Conversations.'' arXiv:2312.06674, 2023.
  \url{https://arxiv.org/abs/2312.06674}. Original-research preprint;
  version and status remain visible.
\item
  Geifman, Y., and El-Yaniv, R. ``SelectiveNet: A Deep Neural Network
  with an Integrated Reject Option.'' \emph{ICML}, 2019, 2151-2159.
  \url{https://proceedings.mlr.press/v97/geifman19a.html}.
\item
  Gama, J., Žliobaitė, I., Bifet, A., Pechenizkiy, M., and Bouchachia,
  A. ``A Survey on Concept Drift Adaptation.'' \emph{ACM Computing
  Surveys} 46(4), Article 44, 2014.
  \url{https://doi.org/10.1145/2523813}. Peer-reviewed scholarly
  synthesis; used here for bounded vocabulary and taxonomy, not original
  empirical or priority claims.
\end{enumerate}

\end{document}